\documentclass[11pt]{article}

\usepackage[preprint]{acl}

\usepackage{times}
\usepackage{latexsym}
\usepackage[T1]{fontenc}
\usepackage[utf8]{inputenc}
\usepackage{microtype}
\usepackage{inconsolata}
\usepackage{graphicx}

\usepackage{amsmath}
\usepackage{amssymb}
\usepackage{amsthm}

\usepackage{booktabs}
\usepackage{multirow}
\usepackage{makecell}
\usepackage{enumitem}
\usepackage{pifont}
\usepackage{placeins}
\usepackage{float}

\usepackage{etoolbox}
\AtBeginEnvironment{table}{\nolinenumbers}
\AtBeginEnvironment{table*}{\nolinenumbers}
\AtBeginEnvironment{figure}{\nolinenumbers}
\AtBeginEnvironment{figure*}{\nolinenumbers}

\newcommand{\cmark}{\ding{51}}
\newcommand{\xmark}{\ding{55}}

\graphicspath{{../../figures/}{../../../figures/}{../../../../figures/}}

\title{Training Leaves Traces: Centered Residual Signatures for Language Model Lineage Verification}

\author{
  \textbf{Aman Singh Thakur}\textsuperscript{1}\textsuperscript{+} \and
  \textbf{Rayan Khoury}\textsuperscript{2} \\\\
  \textsuperscript{1}Amazon, USA \\
  \textsuperscript{2}Massachusetts Institute of Technology, USA \\
  \texttt{amanzing@amazon.com}, \texttt{rkhoury@alum.mit.edu}
}

\begin{document}
\maketitle
{\renewcommand{\thefootnote}{+}\footnotetext{Work does not relate to position at Amazon.}}

\begin{abstract}
Open-weight language models are fine-tuned, quantized, pruned, and merged, yet their provenance is often undocumented. We study data-free white-box lineage verification: \textbf{can weights alone reveal whether two compatible model checkpoints share ancestry?}

Residual training produces a shared identity-aligned component in branch products, so this structure alone cannot establish ancestry. We remove it and compare checkpoint-specific structure across residual blocks, yielding a symmetric lineage score calibrated against independent checkpoints. On residual-MLP and GPT-2 benchmarks, the score separates fine-tuned, LoRA-merged, pruned, and quantized descendants from independent and distilled models (AUROC=1.0), distinguishing weight ancestry from behavioral similarity. Under function-preserving checkpoint laundering experiments, weight-space baselines lose margin or fail; our score remains unchanged and runs 76$\times$ faster than the nearest robust baseline on GPT-2. The projection-pairing signal appears across six language-model families and beyond, and a case study correctly identifies 3 related and 7 unrelated LLaMA-2 public checkpoints. Collectively, these results establish a passive, data-free provenance signal for compatible open-weight language-model checkpoints.
\end{abstract}

\section{Introduction}

\begin{table}[t]
\centering
\small
\begin{tabular}{@{}lp{4.2cm}@{}}
\toprule
Method & Key Limitation \\
\midrule
Crypto Hash & Breaks after any weight change \\
Metadata/Logs & Requires honest distributor \\
Watermarking & Must insert before release \\
Behavioral (CKA, etc.) & Needs data and forward passes \\
Weight Similarity & Breaks under reparameterization \\
Re-Basin & $O(d^3)$ alignment per pair \\
\textbf{Ours} & \emph{Residual architectures only} \\
\bottomrule
\end{tabular}
\caption{Why retroactive lineage verification is hard. Each existing method fails in at least one dimension. Our centered residual signature addresses all key limitations but requires open-weight residual architectures.}
\label{tab:comparison}
\end{table}

\begin{figure*}[t!]
\centering
\includegraphics[width=\textwidth]{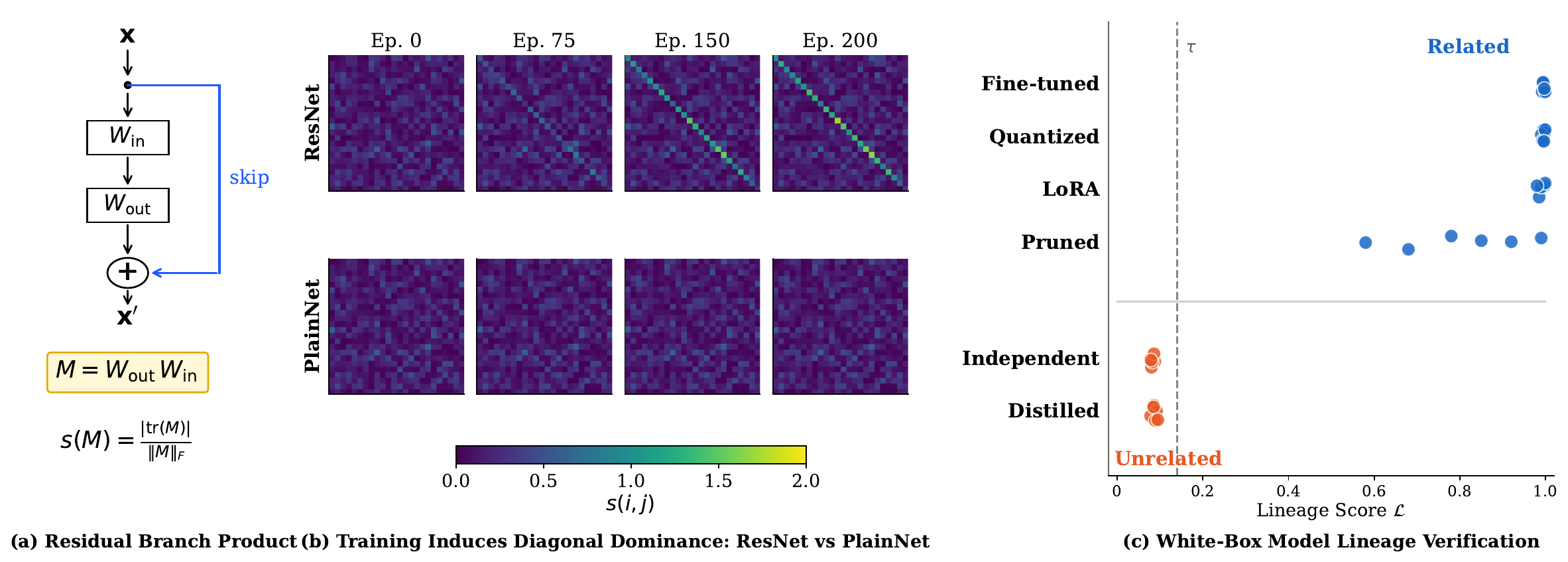}
\caption{Centered residual signatures for lineage verification. (a) The residual branch product $M = W_{\mathrm{out}} W_{\mathrm{in}}$ for a block with skip connection; the normalized trace concentration $s(M)$ measures identity alignment. (b) Training dynamics: score matrices $s(i,j)$ from epoch 0 to 200. ResNet (top row) develops strong identity-aligned structure; PlainNet without skip connections (bottom row) remains random. The structure is learned, not an initialization artifact. (c) Lineage verification: the lineage score $\mathcal{L}$ (Eq.~\ref{eq:lineage-score}) on MLP benchmark (Section~\ref{sec:exp-baselines}) aggregates per-block centered-signature similarity. Descendants (fine-tuned, quantized, pruned, LoRA-merged) score high; non-descendants (independent, distilled) score near zero.}
\label{fig:hero}
\end{figure*}

Open-weight language models are increasingly distributed through supply chains rather than as isolated releases. A base checkpoint may be fine-tuned~\citep{devlin2019bert,ouyang2022rlhf}, quantized~\citep{han2016deep,dettmers2022llmint8}, pruned~\citep{han2015deep}, adapted with low-rank updates, merged~\citep{wortsman2022model,ilharco2023editing}, and redistributed under a new name. These operations create weight descendants, but their ancestry may be undocumented or intentionally obscured. This makes it difficult to distinguish a true descendant from a model that merely behaves similarly.

We study post-hoc white-box model lineage verification. Given reference and suspect checkpoints with compatible residual architectures, the verifier determines from weights alone whether they share weight-level ancestry. Checkpoints are related when they inherit from a common weight ancestor through operations such as fine-tuning, RLHF, pruning, quantization, or LoRA merging. Independently initialized checkpoints are unrelated, even when they share architecture, data, task, or outputs; a distilled student trained from fresh weights is therefore not a weight descendant of its teacher. The task is symmetric: it establishes shared ancestry, not direction of descent.

Existing mechanisms do not solve this setting (Table~\ref{tab:comparison}). Hashes fail after any weight change. Metadata, model cards~\citep{mitchell2019model}, and proof-of-learning records~\citep{jia2021proofoflearning} require trustworthy documentation. Watermarks~\citep{uchida2017embedding,adi2018turning} must be inserted proactively. Behavioral and representation-based methods such as IPGuard~\citep{cao2021ipguard}, CKA~\citep{kornblith2019similarity}, and SVCCA~\citep{raghu2017svcca} require data and forward passes and measure similarity rather than weight inheritance.

Our starting point is a structural property of trained residual networks: correctly paired projections form branch products with concentrated trace. Yet this structure is generic (independently trained models share it), so it cannot establish lineage on its own. We isolate the checkpoint-specific remainder and compare it across blocks, yielding a symmetric lineage score calibrated against independent checkpoints.

The prerequisite trace-concentration signal appears across six language model families (Table~\ref{tab:transformer-family}) and beyond (Appendix~\ref{app:beyond-lm}). Given a trained checkpoint, correctly paired input--output projections within each residual block yield high trace concentration, while mispaired projections score near chance. Hungarian matching on the resulting score matrix recovers corresponding residual block pairings with 100\% accuracy on the canonical MLP path across all tested families, against random-initialization baselines of at most 4\%; alternative paths show more variation (68--100\%; Table~\ref{tab:path-breakdown}). 
On controlled residual-MLP and ~30M-parameter GPT-2 benchmarks, centered residual signatures separate fine-tuned, LoRA-merged, pruned, and quantized descendants from independently trained and distilled models with AUROC=1.0. Simple weight-space baselines also solve clean cases, but under function-preserving checkpoint laundering they lose margin or fail, while our score remains invariant. Even when alignment-based methods recover accuracy, they incur 76$\times$ higher latency on GPT-2. We also run a public LLaMA-2 case study comparing 3 documented descendants against 7 independently trained architectural clones: OpenLLaMA~\citep{geng2023openllama}, Amber~\citep{liu2023llm360}, Baichuan~\citep{yang2023baichuan}, InternLM~\citep{cai2024internlm2}, and Yi~\citep{young2024yi}.

\paragraph{Contributions.}
\begin{enumerate}[leftmargin=*,itemsep=2pt,topsep=2pt]
\item \textbf{Problem Formulation.} We define data-free pairwise weight-lineage verification and introduce centered residual signatures that isolate checkpoint-specific structure.

\item \textbf{Model-level Score.} We construct a symmetric model-level score with empirical-null calibration and algebraic invariance to hidden-unit permutation and reciprocal rescaling.

\item \textbf{Robust Evaluation.} We test robustness to post-training transformations, checkpoint laundering, and distillation, and demonstrate projection-pair recovery across six language model families.
\end{enumerate}

\section{Problem Formulation}\label{sec:background}

\subsection{White-Box Model Lineage Verification}

A verifier receives two checkpoints $A$ and $B$ and determines whether they share weight-level lineage. We assume white-box access and compatible residual architectures (same $L$ and $d$); cross-architecture comparisons are out of scope. The verifier operates on weights alone: no training data, activations, or forward passes.

Checkpoints are \emph{related} if they share a common weight ancestor through common post-training techniques. They are \emph{unrelated} if initialized independently, even when architecture, data, task, or outputs match; a distilled student trained from scratch is unrelated to its teacher, even if it mimics the outputs perfectly. Linear merges induce partial lineage (Appendix~\ref{app:harder-bench}). The verifier returns \textsc{Related} or \textsc{Unrelated}. The score is symmetric, $\mathcal{L}(A,B) = \mathcal{L}(B,A)$; directional attribution requires metadata (Appendix~\ref{app:verification}). The method complements cryptographic signing and watermarking where those were never applied.

\subsection{Related Work}

Table~\ref{tab:comparison} summarizes the comparison.

Cryptographic hashes break after any weight modification. Model cards~\citep{mitchell2019model} require honest distributors and can be falsified when checkpoints are redistributed. Proof-of-learning~\citep{jia2021proofoflearning} requires retained training transcripts that already-released checkpoints lack.

Parameter watermarks~\citep{uchida2017embedding} and backdoor watermarks~\citep{adi2018turning} embed deliberate signals during training. \citet{lukas2022sok} found no surveyed scheme that reliably survived fine-tuning and pruning, and all require proactive insertion: unwatermarked legacy checkpoints cannot be verified.

IPGuard~\citep{cao2021ipguard} extracts inputs near classification boundaries; adversarial frontier stitching~\citep{lemerrer2020adversarial} marks decision surfaces with adversarial examples. CKA~\citep{kornblith2019similarity} and SVCCA~\citep{raghu2017svcca} compare representations on a probe dataset. These methods measure functional or representational similarity. Such similarity does not, by itself, certify weight inheritance, although whether a specific method produces false lineage evidence depends on the model pair and operating threshold. They also require data and forward passes.

Weight cosine, Frobenius distance, and permutation matching~\citep{ainsworth2023git} detect derived checkpoints when the suspect remains close in parameter space, but these are uncalibrated global scalars with no structural account of lineage. HuRef~\citep{zeng2024huref} relies on LLM parameter convergence after pretraining. REEF~\citep{zhang2024reef} needs probe data or activations. MoTHer~\citep{horwitz2025mother} recovers model trees but requires a checkpoint collection rather than a single pair. Our fingerprint is data-free, operates on a single pair against a reference-specific empirical null, and provides per-block similarity scores that track partial overlap (Appendix~\ref{app:harder-bench}).

\citet{park2026idroppedneuralnet} observed that correctly paired projections in trained residual blocks produce trace-concentrated products. Our work builds on this observation: (1) we show this signature appears across six language model families and beyond; (2) we demonstrate that trace concentration alone cannot establish lineage (independently trained models share it), so we decompose the branch product into identity-aligned and checkpoint-specific components, constructing a model-level lineage score validated under common post-training transformations. 
\section{Residual Signatures for Lineage Verification}\label{sec:method}

A residual block computes $x_{\ell+1} = x_\ell + F_\ell(x_\ell)$, where $F_\ell$ is the residual branch. For a branch with $K$ linear layers $W_1, \ldots, W_K$ interleaved with nonlinearities, the \emph{residual branch product} composes the linear factors, dropping nonlinearities:
\begin{equation}\label{eq:branch-product}
M_\ell = W_K W_{K-1} \cdots W_1 \in \mathbb{R}^{d \times d}
\end{equation}
Individual factors $W_k$ are rectangular and do not share an input--output basis; their composed product maps the residual stream back into its own coordinate space, so its diagonal entries compare like with like. The exact factorization is architecture-dependent (Appendix~\ref{app:factorization}, Table~\ref{tab:factorization}).

\subsection{Trace Concentration in Branch Products}\label{sec:trace-concentration}

Under the Frobenius inner product $\langle A, B \rangle = \mathrm{tr}(A^\top B)$, the identity matrix and the traceless matrices span orthogonal subspaces of $\mathbb{R}^{d \times d}$. Any branch product therefore splits into a component along the identity, with coefficient $\langle M_\ell, I \rangle / \langle I, I \rangle = \mathrm{tr}(M_\ell)/d$, and an orthogonal traceless remainder:
\begin{equation}\label{eq:decomp}
M_\ell = \tfrac{\mathrm{tr}(M_\ell)}{d} I + E_\ell
\end{equation}
where $\mathrm{tr}(E_\ell) = 0$. Traces are predominantly negative (Appendix~\ref{app:gpt2-details}); the score uses magnitude.

The \emph{normalized trace concentration}~\citep{park2026idroppedneuralnet} measures this identity alignment:
\begin{equation}\label{eq:dd-score}
s(M) = \frac{|\mathrm{tr}(M)|}{\|M\|_F},
\end{equation}
the fraction of the matrix's energy along the identity direction, up to $\sqrt{d}$. Training moves energy into the identity component: correctly paired branch products score far above unpaired or untrained ones (Figure~\ref{fig:hero}b). The same study shows that the correct-pair score approaches $\sqrt{d}$ as the traceless residual $E$ shrinks, while mismatched projections yield $\mathbb{E}[s] \approx 1/\sqrt{d}$, a margin that grows with dimension.

\begin{table}[t]
\centering
\footnotesize
\begin{tabular}{@{}l@{\hspace{6pt}}r@{\hspace{8pt}}c@{\hspace{8pt}}c@{}}
\toprule
Model & $L$ & Acc (\%) & AUC \\
\midrule
GPT-2~\citep{radford2019gpt2} & 12--48 & 100 & 1.00 \\
BERT~\citep{devlin2019bert} & 12 & 100 & 0.97 \\
LLaMA-2~\citep{touvron2023llama} & 32 & 100 & 1.00 \\
Mistral~\citep{jiang2023mistral} & 32 & 100 & 1.00 \\
Qwen2.5~\citep{yang2024qwen2} & 28 & 100 & 1.00 \\
DeepSeek-R1~\citep{deepseek2025r1} & 32 & 100 & 1.00 \\
\bottomrule
\end{tabular}
\caption{Trace concentration in MLP branch products across language model families. Pair accuracy measures correct within-block projection recovery via Hungarian matching. Random-init baselines: $\le 4\%$. See Appendix~\ref{app:path-breakdown} for per-path results and Appendix~\ref{app:gpt2-details} for GPT-2 scaling.}
\label{tab:transformer-family}
\end{table}

This trace concentration is not specific to MLPs. Table~\ref{tab:transformer-family} shows that the phenomenon appears across six language model families. We extract architecture-aware branch products (Appendix~\ref{app:factorization}) from public checkpoints and apply Hungarian matching on the score matrix $s(i,j)$. Block-pairing accuracy reaches 100\% on the canonical MLP path across all families, against random-initialization baselines of at most 4\%; alternative factorizations show more variation (Table~\ref{tab:path-breakdown}). The extraction transfers across GELU~\citep{hendrycks2016gelu} and SwiGLU~\citep{shazeer2020swiglu} activations with no per-family tuning; only the factorization changes. The phenomenon also transfers to vision (ViT, ResNet) and speech (Whisper) architectures (Appendix~\ref{app:beyond-lm}), so we build lineage signatures on the trace-concentrated structure.

\subsection{A Coupling-Based Account of Identity Alignment}\label{sec:mechanism}

We propose a gradient-coupling account for the observed identity alignment. This is a plausible mechanism, not a proven cause.

The skip connection changes what the branch must learn. A plain layer must transmit the entire representation; a residual branch learns only a correction around an identity path that already transmits everything. This suggests a gradient consequence: $W_{\mathrm{in}}$ and $W_{\mathrm{out}}$ are the two ends of one correction, and the loss reaches both through the same branch. $\nabla W_{\mathrm{out}}$ depends on the branch input, a function of $W_{\mathrm{in}}$; $\nabla W_{\mathrm{in}}$ depends on the upstream gradient, which flows through $W_{\mathrm{out}}$. Their updates may therefore be correlated by construction. In a plain network no such pairing is privileged; every adjacent pair of layers is coupled equally, so no per-block signature can form.

Under this account, the correlation accumulates along the identity direction because the skip means the branch learns a correction, not the full map. Small coordinated corrections to an identity baseline accumulate along the diagonal.

A natural alternative account is dynamical isometry~\citep{pennington2017resurrecting}: if training enforced near-orthogonal block Jacobians $J = I + J_F$, identity alignment would follow as a byproduct of pressure toward $J \approx I$. This predicts trained Jacobians should be more orthogonal than at initialization. We test these accounts with four experiments. First, we train depth-24 residual MLPs on CIFAR-10~\citep{krizhevsky2009cifar} under seven initialization schemes; block-pairing accuracy starts at chance (${\leq}2.1\%$) and reaches 93--100\% after training for every converging scheme (Appendix~\ref{app:mechanism}, Table~\ref{tab:init-ablation}). Orthogonal initialization satisfies dynamical isometry exactly yet yields zero correct pairs at init, confirming the fingerprint requires training.

\begin{figure*}[t]
\centering
\includegraphics[width=\textwidth]{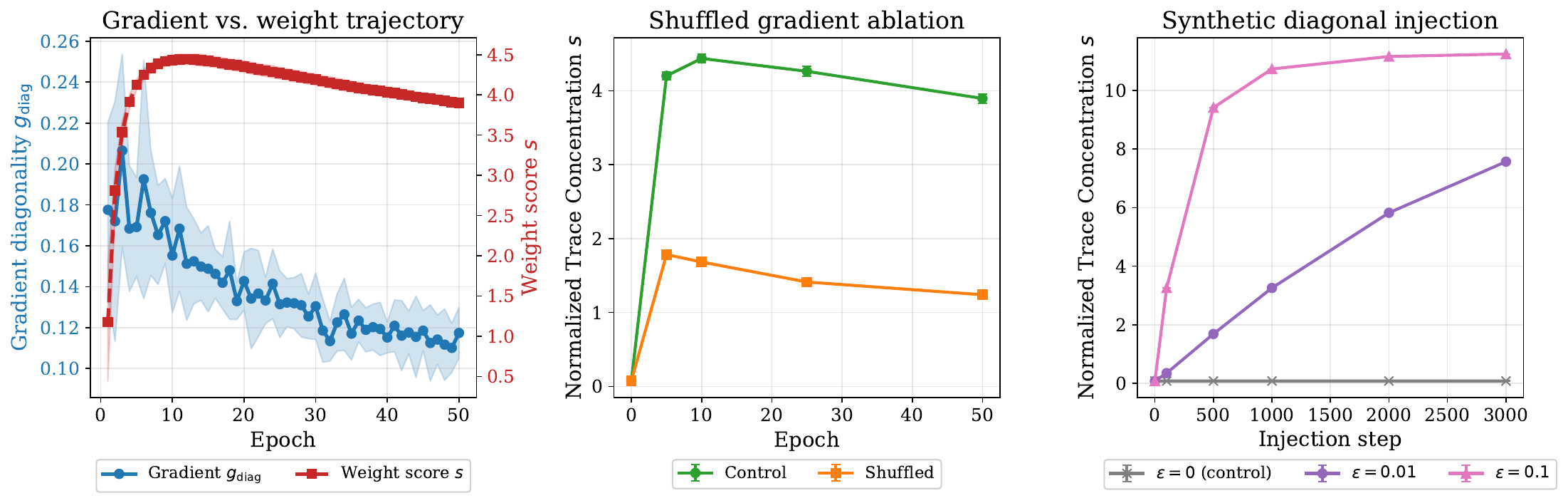}
\caption{Evidence for the gradient-coupling mechanism. (a) Gradient diagonality $g_{\mathrm{diag}}$ (blue) stays flat at ${\sim}0.15$ while the weight score $s$ (red) rises to ${\sim}4.0$: individual gradient updates are not diagonal, but their accumulation is. (b) Shuffling $\nabla W_{\mathrm{out}}$ across blocks (orange) reduces the final score by 68\% compared to control (blue), providing evidence that within-block coupling contributes to the signal. (c) Injecting synthetic diagonal updates $\Delta W = -\varepsilon \cdot e_i^\top$ directly into weights builds the fingerprint from scratch without backpropagation, showing that coordinated identity-aligned updates can generate the observed structure.}
\label{fig:mechanism-triad}
\end{figure*}

Second, we compare ResNet-24 against PlainNet-24 (skip connection removed). ResNet-24 reaches 100\% pair accuracy; PlainNet-24, trained to similar loss, stays at chance (3\%). The skip connection is required.

Third, we measure Jacobian orthogonality across the GPT-2 family. Trained Jacobians are 5--12$\times$ \emph{less} orthogonal than at initialization (GPT-2-small: 0.297 vs 0.025), inconsistent with the isometry hypothesis (Appendix~\ref{app:gpt2-details}).

Fourth, we track gradient diagonality during training (Figure~\ref{fig:mechanism-triad}a). The gradient product $g_{\mathrm{diag}}$ stays flat (${\sim}0.15$), so individual updates are not diagonal, yet the weight score climbs to ${\sim}4.0$. The diagonal structure emerges from accumulated correlated updates. Shuffling gradients across blocks reduces the fingerprint by 68\% (Figure~\ref{fig:mechanism-triad}b); injecting synthetic identity-aligned updates without backpropagation builds it from scratch (Figure~\ref{fig:mechanism-triad}c).

\subsection{Centered Residual Signatures}\label{sec:signatures}

Trace concentration identifies valid trained projection pairings, not ancestry: independently trained models exhibit the same identity-aligned component. The verifier operates on the centered traceless remainder $E_\ell$, using trace concentration only to gate unreliable branches. The identity component is generic; $E_\ell$ is checkpoint-specific, surviving weight-preserving transformations but not reinitialization. On the GPT-2 benchmark, uncentered branch-product cosine similarity shows 0.027 spurious similarity between independent models; centering reduces this to 0.0015, an $18\times$ improvement (Appendix~\ref{app:centering-ablation}).

For each branch product $M_\ell$ we retain only the traceless remainder of decomposition~\eqref{eq:decomp}, normalized to a unit vector:
\begin{equation}\label{eq:centered-fingerprint}
R_\ell = M_\ell - \tfrac{\mathrm{tr}(M_\ell)}{d} \cdot I, \quad
\phi_\ell = \tfrac{\mathrm{vec}(R_\ell)}{\|\mathrm{vec}(R_\ell)\|_2}
\end{equation}
$R_\ell = E_\ell$ from~\eqref{eq:decomp}. The signature $\phi_\ell$ carries checkpoint-specific geometry that weight-preserving transformations inherit and independent training cannot reproduce. Because $M_\ell$ is invariant under hidden permutation and reciprocal rescaling, so is every signature built on it (Section~\ref{sec:exp-laundering}).

Given reference $A$ and suspect $B$ with $L$ blocks each, we form the similarity matrix $G_{ij} = \langle \phi^A_i, \phi^B_j \rangle$, gated by trace concentration (Appendix~\ref{app:verification}), and recover the block correspondence via the Hungarian algorithm~\citep{kuhn1955hungarian}:
\begin{equation}\label{eq:hungarian}
\pi^* = \arg\max_{\pi \in \mathcal{P}_L} \sum_{i} G_{i,\pi(i)}
\end{equation}
For derived checkpoints preserving block order, $\pi^*$ should be the identity; recovering it without metadata confirms that signatures track block identity. Complexity: branch products $O(Ld^2 h)$, similarity matrix $O(L^2 d^2)$, assignment $O(L^3)$.

\subsection{Lineage Score and Verification}\label{sec:lineage-score}

Trace concentration certifies trained residual structure, not ancestry: a verifier built on the trace score alone achieves AUROC$=$0.417 (Appendix~\ref{app:cifar-resnet}, Figure~\ref{fig:lineage-roc}). To distinguish lineage, we compare the centered signatures $\phi_\ell$. The lineage score averages signature similarity over aligned branches:
\begin{equation}\label{eq:lineage-score}
\mathcal{L}(A, B) = \frac{1}{L} \sum_{\ell=1}^{L} \langle \phi^A_\ell, \phi^B_{\pi^*(\ell)} \rangle
\end{equation}
Each term is a cosine similarity, so $\mathcal{L} \in [-1,1]$ with $\mathcal{L}(A,A) = 1$. In practice, all observed scores are non-negative: related checkpoints score near 1, unrelated ones near 0, since independently initialized weights produce uncorrelated $E_\ell$. The score is symmetric.

We calibrate against an empirical null distribution $\mathcal{N}_A = \{\mathcal{L}(A, U_j)\}$ from independently trained models $U_j$. The protocol returns \textsc{Related} if $\mathcal{L}(A,B)$ exceeds the maximum null score, and \textsc{Unrelated} otherwise. Let $\tau = \max_j \mathcal{L}(A, U_j)$:
\begin{equation}\label{eq:verdict}
V(A,B) = \textsc{Related} \;\Leftrightarrow\; \mathcal{L}(A,B) > \tau
\end{equation}
With $n$ null models, this yields a conformal p-value $\leq 1/(n+1)$. The guarantee requires exchangeability: descendants from the same root are not independent samples for calibration purposes, so the effective sample size is governed by the number of independent roots, not the number of pairwise comparisons. We report AUROC as the primary metric, which summarizes discrimination across thresholds without requiring specific false-positive rate claims. Compatibility (same $L$ and $d$) is checked before scoring; the protocol returns \textsc{Incompatible} otherwise. Memory: $O(L^2 + d^2)$ beyond the checkpoints.

\section{Experiments}\label{sec:experiments}


\subsection{Baseline Comparison}\label{sec:exp-baselines}

We compare seven lineage detection methods on two controlled benchmarks with known ancestry: our centered residual signature, weight cosine similarity, aligned Frobenius distance, singular value distance (SVD), SVCCA, CKA, and IPGuard. For the MLP benchmark, we train 2 root models ($L{=}16$, $d{=}48$, input$=$16) for 120 epochs on synthetic regression tasks. From each root, we derive 15 descendants via fine-tuning on the same task (30 epochs), fine-tuning on a different synthetic target function (30 epochs), weight perturbation ($\sigma{=}0.02$), pruning (30\% sparsity), and quantization (8-bit), with 3 variants each. We also train 8 independent models per root using the same architecture but different random seeds, plus 3 distilled students per root ($T{=}2.0$, $\alpha_{\mathrm{CE}}{=}0.5$, 60 epochs) that learn to mimic the root's outputs. This yields 52 pairs: 30 positive (root $\to$ descendant) and 22 negative (16 independent + 6 distilled).

We construct a GPT-2 benchmark using ${\sim}$30M-parameter language models ($L{=}6$, $d{=}384$) trained for 3 epochs on TinyStories~\citep{eldan2023tinystories}. We train 8 root models and derive descendants via continued pretraining on TinyStories (1 epoch), LoRA merge ($r{=}8$, $\alpha{=}16$, 1 epoch), pruning (30/50/70\%), and quantization (8/6-bit). We also train distilled students ($T{=}2.0$, $\alpha_{\mathrm{CE}}{=}\alpha_{\mathrm{KL}}{=}0.5$, 2 epochs). Evaluating on the test-split roots (3 of 8) yields 45 pairs: 21 positive and 24 negative (3 distilled students plus 21 cross-root independent comparisons).

Each (reference, suspect) pair receives a binary label: positive if the suspect inherits weights from the reference, negative otherwise. We report AUROC (threshold-free discrimination) and Gap-$Z$, the standardized margin between positive and negative score distributions:
\begin{equation}\label{eq:gap-z}
\text{Gap-}Z = \frac{\mu_+ - \mu_-}{\sqrt{(\sigma_+^2 + \sigma_-^2)/2}}
\end{equation}
where $\mu_\pm$ and $\sigma_\pm$ are the means and standard deviations of the positive and negative score distributions. Gap-$Z$ measures effect size: values above $+3$ indicate non-overlapping distributions. Because $\sigma_-$ is estimated from few independent roots (3 in GPT-2), so Gap-$Z$ values should be interpreted as approximate.

\begin{table*}[!t]
\centering
\small
\begin{tabular}{@{}llc cc cc@{}}
\toprule
 & & & \multicolumn{2}{c}{MLP (52 pairs)} & \multicolumn{2}{c}{GPT-2 (45 pairs)} \\
\cmidrule(lr){4-5} \cmidrule(lr){6-7}
Method & Type & Data-free & AUROC$\uparrow$ & Gap-$Z$$\uparrow$ & AUROC$\uparrow$ & Gap-$Z$$\uparrow$ \\
\midrule
Centered Residual Signature (ours) & Weight & \cmark & \textbf{1.00} & $+53.0$ & \textbf{1.00} & $\mathbf{+31.0}$ \\
Weight Cosine             & Weight & \cmark & \textbf{1.00} & $\mathbf{+76.3}$ & \textbf{1.00} & $+30.7$ \\
Aligned Frobenius         & Weight & \cmark & \textbf{1.00} & $+72.0$ & \textbf{1.00} & $+3.9$ \\
Singular Value Distance   & Weight & \cmark & \textbf{1.00} & $+7.9$ & 0.73 & $+0.4$ \\
SVCCA~\citep{raghu2017svcca}             & Activation & \xmark & \textbf{1.00} & $+45.4$ & 0.99 & $+3.7$ \\
CKA~\citep{kornblith2019similarity}               & Activation & \xmark & 0.83 & $+8.6$ & 0.86 & $+1.9$ \\
IPGuard~\citep{cao2021ipguard}           & Decision & \xmark & 0.70 & $-3.5$ & 0.91 & $+2.1$ \\
\bottomrule
\end{tabular}
\caption{Baseline comparison for lineage detection. We test whether each method can distinguish checkpoints that share weight ancestry (fine-tuned, pruned, quantized) from independently trained models. MLP benchmark: 52 checkpoint pairs with known ground truth. GPT-2 benchmark: 45 pairs from 30M-parameter language models. AUROC=1.0 means perfect separation; Gap-$Z$>3 indicates non-overlapping score distributions. Weight-space methods (\cmark\ = data-free) match or exceed activation-based methods that require forward passes. See Section~\ref{sec:exp-baselines}.}
\label{tab:baselines}
\end{table*}

Table~\ref{tab:baselines} compares the seven methods across both benchmarks. On MLP, four weight-space methods achieve AUROC$=$1.0: ours, weight cosine, aligned Frobenius, and singular-value distance. SVCCA also achieves perfect discrimination but requires forward passes through both models. CKA and IPGuard underperform (AUROC$=$0.83 and 0.70), with IPGuard's negative Gap-$Z$ indicating it sometimes ranks non-descendants higher than descendants.

On GPT-2, three weight-space methods maintain AUROC$=$1.0: ours, weight cosine, and aligned Frobenius. Singular-value distance degrades (AUROC$=$0.73), suggesting sensitivity to the higher-dimensional weight matrices in transformer MLPs. Among activation-based methods, SVCCA remains strong (0.99) while CKA stays weak (0.86). IPGuard improves on GPT-2 (0.91) but still lags behind weight-space approaches.

We validate on public checkpoints (Table~\ref{tab:real-llm-baselines}): comparing LLaMA-2-7B-base against 3 documented derivatives and 7 independently trained models that share identical architecture (32 layers, $d{=}4096$, intermediate$=$11008, SwiGLU). The derivatives produce high scores ($\mathcal{L}$: 0.995, 0.996, 0.336) while all 7 independent models score $|\mathcal{L}| < 5 \times 10^{-5}$.

\begin{table}[t]
\centering\small
\begin{tabular}{@{}lr@{}}
\toprule
Suspect & Score ($\mathcal{L}$) \\
\midrule
\multicolumn{2}{@{}l}{\textit{Descendants}} \\
\quad Llama-2-7B-chat~\citep{touvron2023llama}   & $0.995$ \\
\quad Vicuna-7B~\citep{chiang2023vicuna}         & $0.996$ \\
\quad CodeLlama-7B~\citep{roziere2023code}      & $0.336$ \\
\midrule
\multicolumn{2}{@{}l}{\textit{Non-descendants}} \\
\quad OpenLLaMA-7B~\citep{geng2023openllama}      & $4\mathrm{e}{-}4$ \\
\quad OpenLLaMA-7B-v2~\citep{geng2023openllama}   & $-2\mathrm{e}{-}5$ \\
\quad Amber~\citep{liu2023llm360}                 & $-3\mathrm{e}{-}5$ \\
\quad Baichuan-7B~\citep{yang2023baichuan}        & $-5\mathrm{e}{-}5$ \\
\quad Baichuan2-7B~\citep{yang2023baichuan}       & $-5\mathrm{e}{-}6$ \\
\quad InternLM-7B~\citep{cai2024internlm2}        & $2\mathrm{e}{-}5$ \\
\quad Yi-6B~\citep{young2024yi}                   & $2\mathrm{e}{-}5$ \\
\bottomrule
\end{tabular}
\caption{Public checkpoint case study (LLaMA-2 family). Descendants score $\mathcal{L} \geq 0.3$; all 7 independent models score $|\mathcal{L}| < 5 \times 10^{-5}$.}
\label{tab:real-llm-baselines}
\end{table}

\subsection{Robustness Under Function-Preserving Checkpoint Laundering}\label{sec:exp-laundering}

Table~\ref{tab:baselines} shows that all weight-space baselines achieve strong results on clean benchmarks. We next evaluate a more stringent condition: \emph{function-preserving laundering}, where an adversary disguises a stolen model by rearranging its internal structure (shuffling neurons or rescaling weights) without changing what the model computes. We evaluate all weight-space baselines plus Re-Basin+scale~\citep{ainsworth2023git}, which recovers alignment via $L$ Hungarian assignments per model pair.

To generate reliably laundered checkpoints, we use two function-preserving operations. Per-block permutation (P) reorders hidden units ($W_{\mathrm{in}} \leftarrow P W_{\mathrm{in}}$, $W_{\mathrm{out}} \leftarrow W_{\mathrm{out}} P^\top$); reciprocal rescaling (D) scales them ($W_{\mathrm{in}} \leftarrow D W_{\mathrm{in}}$, $W_{\mathrm{out}} \leftarrow W_{\mathrm{out}} D^{-1}$). These are the natural function-preserving symmetries of MLP blocks: P exploits neuron-ordering equivalence; D exploits positive homogeneity of ReLU-family activations. Orthogonal rotation $Q$ of the residual stream ($M \mapsto QMQ^\top$) would also cancel in the branch product, but is not function-preserving for models with learned LayerNorm or RMSNorm parameters (Limitations). Re-Basin+scale can recover alignment under P and D, but at higher computational cost: it solves $L$ Hungarian assignments per pair ($O(Ld^3)$), while our method computes branch products directly ($O(Ld^2 h)$). All laundered variants pass a function-preservation gate: the model's outputs change by at most $10^{-4}$.

We take the 52-pair MLP benchmark ($L{=}16$, $d{=}48$) and launder six variants: none (baseline), P (random permutation per block), D$_\mathrm{m}$ (mild rescaling, log-uniform $[0.5,2]$), D$_\mathrm{s}$ (strong rescaling, log-uniform $[0.1,10]$), PD (permutation then rescaling), and PDFT (PD followed by 5 epochs fine-tuning). We also run P on the GPT-2 benchmark but cannot run D variants because GPT-2 uses GELU activation, which is not positively homogeneous ($\mathrm{GELU}(\alpha x) \neq \alpha \cdot \mathrm{GELU}(x)$), so rescaling is not function-preserving.

\begin{table}[t]
\centering
\small
\setlength{\tabcolsep}{3pt}
\begin{tabular}{@{}llccccc@{}}
\toprule
& & \multicolumn{5}{c}{AUROC} \\
\cmidrule(lr){3-7}
Cond. & Bench & \textbf{Ours} & Re-Basin & Al.~Frob & SVD & W.~Cos \\
\midrule
\multirow{2}{*}{P}
  & MLP & \textbf{1.0} & \textbf{1.0} & 0.50 & \textbf{1.0} & 0.86 \\
  & GPT-2 & \textbf{1.0} & \textbf{1.0} & \textbf{1.0} & 0.76 & \textbf{1.0}$^\dagger$ \\
\midrule
D$_\mathrm{m}$ & MLP & \textbf{1.0} & \textbf{1.0} & \textbf{1.0} & 0.0 & \textbf{1.0} \\
D$_\mathrm{s}$ & MLP & \textbf{1.0} & \textbf{1.0} & 0.0 & 0.0 & \textbf{1.0} \\
PD & MLP & \textbf{1.0} & \textbf{1.0} & 0.0 & 0.0 & 0.80 \\
PDFT & MLP & \textbf{1.0} & \textbf{1.0} & 0.0 & 0.0 & 0.80 \\
\midrule
\multirow{2}{*}{Lat.}
  & MLP & \textbf{0.4} & 0.8 & 1.3 & 2.6 & 1.5 \\
  & GPT-2 & \textbf{5} & 388 & 25 & 1438 & 50 \\
\bottomrule
\end{tabular}
\caption{AUROC under function-preserving laundering. P$=$permutation, D$_\mathrm{m/s}$$=$mild/strong rescaling, Lat.$=$latency (ms). Our method and Re-Basin maintain AUROC$=$1.0 across all conditions; raw baselines collapse. Ours is 2$\times$ faster on MLP, 76$\times$ faster on GPT-2. $^\dagger$Weight cosine scores collapse 97\% (see Appendix~\ref{app:laundering-gapz}).}
\label{tab:laundering}
\end{table}

Table~\ref{tab:laundering} compares methods under laundering. Our signature and Re-Basin+scale both maintain AUROC$=$1.0 across all conditions, while raw baselines collapse: aligned Frobenius drops to 0.50 under permutation on MLP and 0.0 under strong rescaling; SVD fails rescaling entirely; weight cosine degrades to 0.80 under PD. Aligned Frobenius retains AUROC$=$1.0 on GPT-2 under permutation, but this is an artifact of the block-level Hungarian assignment: with only 6 blocks (vs.\ 16 in MLP), per-block norm statistics remain distinctive enough to recover correct layer correspondence even when within-block similarity is destroyed. The Gap-$Z$ margin still collapses from $+72$ to $+3.9$ (Appendix~\ref{app:laundering-gapz}). The key differentiator is computational cost. Re-Basin+scale solves $L$ Hungarian assignments per pair ($O(Ld^3)$ complexity), while our method computes branch products directly ($O(Ld^2 h)$). This yields 2$\times$ speedup on MLP (0.4ms vs 0.8ms) and 76$\times$ speedup on GPT-2 (5ms vs 388ms; see Appendix~\ref{app:latency} for detailed statistics). Our method achieves invariance algebraically, with no alignment search.

AUROC alone understates the difference: weight cosine's Gap-$Z$ collapses from $+76$ to $+2$ under permutation (Appendix~\ref{app:laundering-gapz}), while our signature maintains Gap-$Z \approx +53$ across all conditions.

We replicate on public checkpoints from four 7B-scale model families (Table~\ref{tab:laundering-llm}): LLaMA-2~\citep{touvron2023llama}, LLaMA-3~\citep{grattafiori2024llama3}, Mistral~\citep{jiang2023mistral}, and Qwen2.5~\citep{yang2024qwen2}. For each base model and its fine-tuned derivative, we shuffle hidden units in the derivative's MLP layers and measure how much the lineage scores change. Our score is unaffected ($\Delta\mathcal{L} < 10^{-7}$); weight cosine loses ${\sim}95\%$ of its signal.

\begin{table}[t]
\centering
\small
\begin{tabular}{@{}llcc@{}}
\toprule
Base & Suspect & Relation & $\Delta$W.cos \\
\midrule
LLaMA-2 7B & Chat & derivative & 0.993 \\
LLaMA-3 8B & Instruction-tuned & derivative & 0.949 \\
Mistral 7B & Instruction-tuned & derivative & 0.934 \\
Qwen2.5 7B & Instruction-tuned & derivative & 0.948 \\
\midrule
LLaMA-2 7B & OpenLLaMA & independent & (0.089)$^\dagger$ \\
LLaMA-3 8B & Mistral 7B & independent & (0.158)$^\dagger$ \\
\bottomrule
\end{tabular}
\caption{Permutation laundering on public language model derivatives. $\Delta$W.cos = score before $-$ score after for derivatives. Our lineage score $\Delta\mathcal{L} < 10^{-7}$ (invariant); weight cosine loses ${>}93\%$ of its signal in all derivative cases. $^\dagger$Raw W.cos score (independently trained).}
\label{tab:laundering-llm}
\end{table}

\subsection{Robustness to Post-Training Techniques}\label{sec:exp-posttraining}

Section~\ref{sec:exp-baselines} showed that all weight-space baselines achieve AUROC$=$1.0 on clean benchmarks. We evaluate robustness to common post-training transformations that modify weights while preserving model behavior.

For MLPs, we train 3 root models ($L{=}24$, $d{=}48$) for 120 epochs on synthetic regression tasks. Per root we generate 25 descendants via fine-tuning, weight perturbation, pruning (10--85\% sparsity), and quantization (16--256 levels), plus 28 unrelated checkpoints (independent same-task, different-task, random-init, and distilled students). This yields 159 pairs: 75 positive and 84 negative. Figure~\ref{fig:hero}c visualizes MLP results: descendants cluster near $\mathcal{L}{=}1$ while non-descendants fall below the calibrated threshold. Even at 85\% sparsity, the minimum descendant score (0.58) remains ${\sim}3\times$ above the maximum non-descendant score (0.20).

\begin{table}[t]
\centering\small
\setlength{\tabcolsep}{4pt}
\begin{tabular}{@{}lccc@{}}
\toprule
Transformation & $n$ & Mean $\mathcal{L}$ & Min $\mathcal{L}$ \\
\midrule
\multicolumn{4}{@{}l}{\textit{Descendants}} \\
\quad Quantized (INT8/6) & 6 & 0.999 & 0.999 \\
\quad LoRA merge (rank-8) & 3 & 0.998 & 0.996 \\
\quad Fine-tuned (1 epoch) & 3 & 0.980 & 0.980 \\
\quad Pruned (30--70\%) & 9 & 0.937 & 0.855 \\
\midrule
\multicolumn{4}{@{}l}{\textit{Non-descendants}} \\
\quad Distilled student & 3 & 0.002 & 0.001 \\
\quad Independent & 21 & 0.003 & 0.000 \\
\bottomrule
\end{tabular}
\caption{Lineage scores under post-training transformations (GPT-2 benchmark from Section~\ref{sec:exp-baselines}, 3 test roots). The minimum descendant score (0.855 at 70\% sparsity) exceeds the maximum non-descendant score (0.004).}
\label{tab:gpt2-survival}
\end{table}

Table~\ref{tab:gpt2-survival} reports GPT-2 lineage scores on the same benchmark (Section~\ref{sec:exp-baselines}), broken down by transformation type. Quantization and LoRA preserve the signal ($\mathcal{L}{\geq}0.996$); fine-tuning and pruning degrade it but remain well above the null ($\mathcal{L}{\geq}0.855$). Both distilled students and independently trained models score near zero. The $200\times$ gap between minimum descendant and maximum non-descendant scores yields AUROC$=$1.0 with perfect separation.

Behavioral similarity does not imply weight ancestry: distillation increases teacher--student agreement (+1.4 points top-1) while producing no ancestry signal ($\mathcal{L} \approx 0.002$), confirming our method detects shared parameters, not behavioral mimicry (Appendix~\ref{app:gpt2-benchmark}).

\section{Conclusion}

We introduced a data-free lineage detection score derived from checkpoint-specific structure in residual branch products. Across residual-MLP and GPT-2 benchmarks, the score separated fine-tuned, LoRA-merged, pruned, and quantized descendants from independently trained and distilled models, while distinguishing weight ancestry from behavioral similarity. It remained stable under function-preserving checkpoint laundering and matched the nearest robust baseline at lower latency. Projection-pairing results across six language-model families and a public LLaMA-2 case study support the approach. The method is limited to compatible white-box residual checkpoints and does not establish ownership or direction of descent. It enables provenance auditing for open-weight language-model supply chains.

\section*{Ethical Considerations}

We present a method for detecting shared weight ancestry between neural network checkpoints. Potential applications include model provenance verification and supply-chain auditing. We acknowledge several ethical considerations:

The method produces a similarity score, not a legal determination. Scores consistent with documented relationships do not independently prove provenance, ownership, or wrongdoing. Results should be combined with metadata, release records, and human review before drawing conclusions about model origins.

Limited calibration data (few independent roots) may produce spuriously precise thresholds. Users should understand that formal hypothesis testing requires sufficient exchangeable samples, which may not be available for rare architectures.

The method could potentially be used to identify undocumented model relationships, which may conflict with developers' reasonable expectations of anonymity. The white-box access requirement limits covert use, but authorized access does not automatically imply consent to provenance analysis.

The same method that enables legitimate supply-chain auditing could be misused for unfounded infringement claims or competitive intelligence. We encourage responsible deployment with appropriate evidentiary standards.

\section*{Limitations}\label{sec:limitations}

Our method needs white-box access to both checkpoints, so API-only models cannot be verified. It also only works for residual architectures: plain feedforward networks, RNNs, and state-space models are out of scope. Comparisons are further limited to models with matching depth and hidden dimension, so cross-architecture verification (e.g., LLaMA vs.\ GPT-2) is not supported.

The lineage score is symmetric, so without external metadata we cannot tell which checkpoint is the ancestor. The score decays monotonically with genealogical distance, but the method compares single pairs and does not reconstruct multi-hop ancestry or family trees. Under partial inheritance (layer grafts, linear merges), the score reflects the fraction of shared blocks but not which blocks were inherited.

The fingerprint survives common post-training modifications, but weakens as transformations grow more aggressive. Heavy pruning drops the signal to $\mathcal{L}=0.58$ at 85\% sparsity, and extensive continued pretraining (CodeLlama) lowers it to $\mathcal{L}=0.336$, though both remain detectable above the null. In the evaluated MLP suppression attack, reaching the null threshold ($\mathcal{L} \approx 0.084$) costs +1.5\% utility loss; driving the score reliably below null costs +12\%. This result does not establish robustness to other attacks, architectures, utility measures, or optimization budgets.

The signature is invariant to hidden-unit permutation and reciprocal rescaling within MLP blocks, but not to orthogonal rotation $Q$ of the residual stream: $M \mapsto QMQ^\top$ preserves trace but transforms the centered remainder, destroying cross-checkpoint similarity. For LayerNorm architectures (GPT-2, BERT), such rotation is not function-preserving because learned element-wise $\gamma$ and $\beta$ parameters do not commute with arbitrary $Q$. For RMSNorm architectures with learned per-dimension scale (LLaMA, Mistral), the same obstruction applies unless $\gamma$ is constant across dimensions, which is not standard practice. An architecture using only unparameterized normalization (e.g., pure RMS without learned scale) would be vulnerable to $Q$-rotation laundering; we are not aware of widely deployed models in this category.

Verification requires calibrating a null distribution from independently trained models of the same architecture, and the threshold may need adjustment across architecture families. Our controlled benchmarks demonstrate perfect AUROC discrimination, but the effective sample sizes for formal hypothesis testing are limited: the GPT-2-style benchmark has 8 roots total. The LLaMA-2 case study includes 7 independent models (OpenLLaMA v1/v2, Amber, Baichuan v1/v2, InternLM, Yi), providing stronger calibration for that architecture family, but other families remain undertested. Real-world validation covers a limited set of public checkpoints, so broader ecosystem coverage is untested. SwiGLU architectures (Qwen2.5, DeepSeek-R1) reach only 80\% block-pairing accuracy and need joint factorization to recover sub-paths reliably.

\section*{Use of AI Assistants}

The authors bear full responsibility for all scientific claims, experimental results, and analysis in this work. During manuscript preparation, we used Claude to support literature searches, experimental design, code development, and the drafting and editing of text. All AI-assisted material was reviewed, verified, and revised by the authors.

\bibliography{../../custom}

\appendix
\section{Reproducibility Statement}\label{app:reproducibility}

All results in this paper are produced by deterministic scripts with fixed seeds. The 52-pair MLP lineage benchmark (Tables~\ref{tab:survival}, \ref{tab:baselines}) uses depth-16 MLPs (in\_dim=16, hidden=48) and is constructed from $2$ trained reference checkpoints, each paired with $5$ related-checkpoint kinds $\times 3$ derivative seeds $=30$ truly-related pairs (fine-tune, fine-tune to a new target, $\sigma_{\text{rel}} \in [0.01,0.15]$ Gaussian noise, $10$--$85\%$ magnitude pruning, and $16$--$256$-level uniform quantization), plus $16$ different-seed same-task pairs and $6$ distilled-student pairs as unrelated checkpoints. The expanded 159-pair set (Figure~\ref{fig:hero}c) uses $3$ depth-24 MLP reference models: each reference is paired with $25$ related checkpoints ($5$ transformation types $\times$ $5$ variations) and $28$ unrelated checkpoints ($15$ independent same-task $+$ $5$ independent different-task $+$ $5$ random-init $+$ $3$ distilled), yielding $3 \times 25 = 75$ related pairs and $3 \times 28 = 84$ unrelated pairs ($75 + 84 = 159$ total). These 84 unrelated pairs share only 3 reference model roots; the effective sample size for null calibration is limited by the number of independent roots, not the number of pairs.

The GPT-2-Small-Lite benchmark (Table~\ref{tab:gpt2-survival}) uses 8 transformer roots with GPT-2 architecture ($L{=}6$ layers, $d{=}384$, $d_{\mathrm{ff}}{=}1536$, 6 heads, ${\sim}$30M parameters), trained on TinyStories~\citep{eldan2023tinystories} (50K samples, 3 epochs, batch size 8, lr$=$3e-4, cosine schedule, fp16, gradient checkpointing). Roots are allocated to calibration ($n{=}2$), development ($n{=}3$), and test ($n{=}3$) splits; Table~\ref{tab:gpt2-survival} reports results only on the 3 held-out test roots (21 descendant pairs, 3 distilled pairs, 21 cross-root pairs $=$ 45 total). Each root generates 7 descendants: 1 fine-tuned (1 epoch continued pretraining, lr$=$1e-4), 1 LoRA-merged (rank-8, $\alpha{=}16$, 1 epoch, lr$=$1e-4), 3 pruned (magnitude pruning at 30/50/70\% sparsity), and 2 quantized (uniform quantization at 256/64 levels, i.e., INT8/INT6). We also train 8 distilled students (one per root) are trained from scratch using knowledge distillation ($T{=}2.0$, $\alpha_{\mathrm{CE}}{=}0.5$, $\alpha_{\mathrm{KL}}{=}0.5$, 2 epochs). Quality metrics for distilled students include top-1 agreement, KL divergence, and perplexity ratio, computed on a held-out validation set.

Real-architecture pairing (Table~\ref{tab:transformer-family}) loads public HuggingFace checkpoints. Code, configuration files, baseline implementations (aligned Frobenius, singular-value distance, weight cosine, CKA, SVCCA, IPGuard regression analogue), the per-pair score JSONs, and the plotting scripts will be released upon acceptance.

\paragraph{Hardware and software.} All experiments were run on a single NVIDIA L4 GPU (24GB VRAM) with an AMD EPYC 7R13 CPU. Software: PyTorch, NumPy, SciPy, and Hugging Face Transformers. Global seeds (\texttt{torch.manual\_seed(0)}, \texttt{np.random.seed(0)}) are set at the start of each benchmark script; per-model seeds are derived deterministically from root and variant indices.

\paragraph{Data sources.} TinyStories~\citep{eldan2023tinystories} is loaded via Hugging Face Datasets (\texttt{roneneldan/TinyStories}). CIFAR-10~\citep{krizhevsky2009cifar} uses \texttt{torchvision.datasets}. Public checkpoints (GPT-2, LLaMA-2, Mistral, Qwen) are loaded from their official Hugging Face repositories.

\subsection{Latency Statistics}\label{app:latency}

Table~\ref{tab:gpt2-latency} reports per-method latency on the GPT-2 benchmark (20 pairs, GPU). Our method is 76$\times$ faster than Re-Basin+scale and 283$\times$ faster than SVD, while achieving identical AUROC under laundering.

\begin{table}[h]
\centering
\small
\setlength{\tabcolsep}{4pt}
\begin{tabular}{@{}lrrrr@{}}
\toprule
Method & Mean & Std & Min & Max \\
\midrule
Ours & \textbf{5.1} & 2.3 & 4.4 & 15.1 \\
Re-Basin+scale & 387.9 & 10.9 & 369.6 & 424.6 \\
Aligned Frob.\ & 25.0 & 1.3 & 23.6 & 29.2 \\
SVD & 1438.2 & 35.5 & 1417.7 & 1585.1 \\
Weight Cos.\ & 50.1 & 7.7 & 46.2 & 83.1 \\
\bottomrule
\end{tabular}
\caption{Per-method latency (ms) on GPT-2 benchmark. Ours achieves 76$\times$ speedup over Re-Basin+scale.}
\label{tab:gpt2-latency}
\end{table}

\subsection{Architecture-Aware Factorization}\label{app:factorization}

Table~\ref{tab:factorization} lists the architecture-aware residual branch products used throughout this work.

\begin{table*}[t]
\centering
\small
\begin{tabular}{@{}lll@{}}
\toprule
Architecture & Residual branch $F(x)$ & Product $M$ \\
\midrule
Transformer MLP / BasicBlock & $W_2 \sigma(W_1 x)$ & $W_2 W_1$ \\
Attention V/O path & $W_O \mathrm{Attn}(x) W_V x$ & $W_O W_V$ \\
Attention Q/K path & bilinear $x^\top W_Q W_K^\top x$ & $W_Q W_K^\top$ \\
Bottleneck ResNet & $W_3 \sigma(W_2 \sigma(W_1 x))$ & $W_3 W_2 W_1$ \\
SwiGLU MLP & $W_{\mathrm{down}} [\sigma(W_{\mathrm{gate}} x) \odot W_{\mathrm{up}} x]$ & $W_{\mathrm{down}} W_{\mathrm{up}}$ \\
\bottomrule
\end{tabular}
\caption{Architecture-aware residual branch products. Each $M$ composes the linear maps of one branch (dropping nonlinearities) so it maps the residual stream to itself. Factorization must match the architecture; incomplete products destroy the signal.}
\label{tab:factorization}
\end{table*}

On ImageNet-pretrained ResNet-50/101/152 (torchvision), we compare naive two-layer factorization ($W_3 W_1$, skipping the middle conv) against the correct three-layer product ($W_3 W_2 W_1$) on layer3 bottleneck blocks ($n{=}5/22/35$ blocks respectively). Naive $W_3 W_1$ achieves chance-level accuracy; the correct triple product recovers:
\begin{itemize}[nosep]
\item ResNet-50/layer3 ($n{=}5$): 100\%, AUC 1.000
\item ResNet-101/layer3 ($n{=}22$): 100\%, AUC 0.995
\item ResNet-152/layer3 ($n{=}35$): 91\%, AUC 0.964
\end{itemize}

\subsection{Per-Path Pairing Accuracy}\label{app:path-breakdown}

Table~\ref{tab:transformer-family} reports only the canonical MLP path ($W_{\mathrm{down}} W_{\mathrm{up}}$). Table~\ref{tab:path-breakdown} provides the full per-path breakdown across all tested factorizations.

\begin{table*}[t]
\centering
\small
\setlength{\tabcolsep}{4pt}
\begin{tabular}{@{}llccccc@{}}
\toprule
& & \multicolumn{3}{c}{MLP Paths} & \multicolumn{2}{c}{Attention Paths} \\
\cmidrule(lr){3-5} \cmidrule(lr){6-7}
Model & $L$ & down$\times$up & down$\times$gate & joint & $W_O W_V$ & $W_Q W_K^\top$ \\
\midrule
\multicolumn{2}{@{}l}{\textit{Architecture}} & SwiGLU / GELU & SwiGLU & SwiGLU & Attention & Attention \\
\midrule
GPT-2 (124M--1.5B) & 12--48 & \textbf{100\%} & --- & --- & --- & --- \\
BERT-base & 12 & \textbf{100\%} & --- & --- & \textbf{100\%} & \textbf{100\%} \\
LLaMA-2-7B-chat & 32 & \textbf{100\%} & \textbf{100\%} & \textbf{100\%} & \textbf{100\%} & \textbf{100\%} \\
Mistral-7B & 32 & \textbf{100\%} & \textbf{100\%} & \textbf{100\%} & \textbf{100\%} & \textbf{100\%} \\
Qwen2.5-7B & 28 & \textbf{100\%} & 68\% & \textbf{100\%} & \textbf{100\%} & \textbf{100\%} \\
DeepSeek-R1-Distill & 32 & \textbf{100\%} & 84\% & \textbf{100\%} & \textbf{100\%} & \textbf{100\%} \\
\midrule
Random-init baseline & --- & 3--4\% & 3--4\% & 4--7\% & 3--9\% & 0--6\% \\
\bottomrule
\end{tabular}
\caption{Per-path pairing accuracy across language model families. The canonical MLP path (down$\times$up) achieves 100\% across all models. The alternative SwiGLU factorization (down$\times$gate) shows weaker signal in some architectures (Qwen: 68\%, DeepSeek: 84\%), likely because the gating path carries less of the residual correction. GPT-2 and BERT use GELU activation with only one MLP factorization ($W_2 W_1$). Dashes indicate paths not applicable to the architecture. See Table~\ref{tab:factorization} for the exact product $M$ corresponding to each path.}
\label{tab:path-breakdown}
\end{table*}

\section{Verification Protocol Details}\label{app:verification}

\subsection{Gated Branch Score}

Branch similarity is reliable only when both branches exhibit strong trace concentration. We gate the cosine similarity:
\begin{equation}
G(\phi^A_\ell, \phi^B_m) = \langle \phi^A_\ell, \phi^B_m \rangle \cdot \min\!\left(\tfrac{s(M^A_\ell)}{\tau_s}, \tfrac{s(M^B_m)}{\tau_s}, 1\right)
\end{equation}
where $\tau_s$ is the minimum trace concentration across reference branches.

\subsection{Statistical Calibration}

We calibrate against a null distribution $\mathcal{N}_A = \{\mathcal{L}(A, U_j) : U_j \notin \mathcal{D}(A)\}$ of scores between reference $A$ and independently trained models. Given $n$ null scores, the empirical threshold $\tau = \max_j \mathcal{L}(A, U_j)$ yields a conformal p-value:
\begin{equation}
p(A,B) = \frac{|\{j : \mathcal{L}(A, U_j) \geq \mathcal{L}(A,B)\}| + 1}{n + 1}
\end{equation}
With $n$ null models, the minimum achievable p-value is $1/(n+1)$. We therefore report AUROC as the primary metric, which aggregates discrimination across all thresholds without requiring specific FPR claims.

\paragraph{Calibration power by benchmark.} The MLP benchmark has 84 unrelated pairs from 3 independent roots; the effective sample size for calibration is 3 roots, not 84 pairs, since descendants from the same root are not independent samples. The GPT-2-Small-Lite benchmark uses 8 roots total, with 3 held out for testing; each test root is compared against 7 independent roots, but the effective sample size for formal hypothesis testing remains limited. We report AUROC as a discrimination summary rather than threshold-based verification claims with specific false-positive rate guarantees. The LLaMA-2 case study includes 7 independently trained architectural clones (OpenLLaMA v1/v2, Amber, Baichuan v1/v2, InternLM, Yi), all with identical architecture (32 layers, $d{=}4096$, intermediate$=$11008). All 7 score $|\mathcal{L}| < 5 \times 10^{-5}$ against the LLaMA-2-7B reference, while descendants score $\mathcal{L} \geq 0.3$, providing meaningful null calibration for this architecture family.

\begin{table}[h]
\centering
\small
\setlength{\tabcolsep}{3pt}
\begin{tabular}{@{}lccc@{}}
\toprule
Transformation & Mean & Min & $>$null \\
\midrule
FT / Quant / Noise & 0.99 & 0.97 & 45/45 \\
FT (diff.\ target) & 0.94 & 0.84 & 15/15 \\
Pruning (10--85\%) & 0.81 & 0.58 & 15/15 \\
\midrule
Indep.\ / Distilled & 0.08 & 0.01 & - \\
\bottomrule
\end{tabular}
\caption{Lineage score $\mathcal{L}$ survival under post-training modification ($n{=}75$ related, $n{=}84$ unrelated). All related exceed $\max \mathcal{L}_{\text{null}} = 0.20$; AUROC$=$1.0.}
\label{tab:survival}
\end{table}

\section{Mechanism Validation}\label{app:mechanism}

\subsection{Initialization Ablation}\label{app:init-ablation}

Table~\ref{tab:init-ablation} shows pair accuracy before and after training across seven initialization schemes on depth-24 residual MLPs (3 seeds, 200 epochs). All schemes show chance-level accuracy (${\leq}2.1\%$) at initialization, rising to $93$--$100\%$ after training. Orthogonal initialization provides dynamical isometry at init yet shows zero correct pairs before training.

\begin{table}[h]
\centering
\small
\begin{tabular}{@{}lccc@{}}
\toprule
Init scheme & Untrained & Trained & AUROC \\
\midrule
Orthogonal       & $0.0\%$  & $100\%$ & $0.947$ \\
Kaiming-normal   & $2.1\%$  & $97\%$  & $0.981$ \\
Kaiming-uniform  & $2.1\%$  & $98.6\%$ & $0.98$ \\
Xavier-normal    & $2.1\%$  & $100\%$ & $0.990$ \\
Xavier-uniform   & $2.1\%$  & $100\%$ & $0.99$ \\
Uniform          & $2.1\%$  & $93.8\%$ & --- \\
Gaussian $\sigma{=}0.02$ & $2.1\%$ & $13\%$ & $0.671$ \\
\bottomrule
\end{tabular}
\caption{Initialization ablation on depth-24 residual MLPs. All schemes show chance-level accuracy before training. The Gaussian $\sigma{=}0.02$ case fails because blocks collapse to near-zero contribution.}
\label{tab:init-ablation}
\end{table}

\subsection{GPT-2 Scaling, Trace, and Jacobian Analysis}\label{app:gpt2-details}

Table~\ref{tab:jacobian-orthogonality} reports the normalized Jacobian deviation $\delta_J^{\mathrm{norm}} = \|J^\top J/\|J\|_F^2 - I/d\|_F$ across GPT-2 scales. This metric is scale-invariant: any scaled orthogonal matrix scores 0, and larger values indicate non-uniform singular values.

\begin{table}[h]
\centering
\small
\setlength{\tabcolsep}{3pt}
\begin{tabular}{@{}lcccr@{}}
\toprule
Model & $d$ & Pretrained & Rand-init & Ratio \\
\midrule
GPT-2-small  & 768  & 0.297 & 0.025 & $11.7\times$ \\
GPT-2-medium & 1024 & 0.242 & 0.026 & $9.4\times$ \\
GPT-2-large  & 1280 & 0.155 & 0.025 & $6.2\times$ \\
GPT-2-XL     & 1600 & 0.122 & 0.024 & $5.1\times$ \\
\bottomrule
\end{tabular}
\caption{Jacobian orthogonality ($\delta_J^{\mathrm{norm}}$) across GPT-2 scales. Pretrained models are $5$--$12\times$ less orthogonal than at random init, inconsistent with the hypothesis that trace concentration results from blocks approaching isometry.}
\label{tab:jacobian-orthogonality}
\end{table}

\begin{figure*}[t]
\centering
\includegraphics[width=\textwidth]{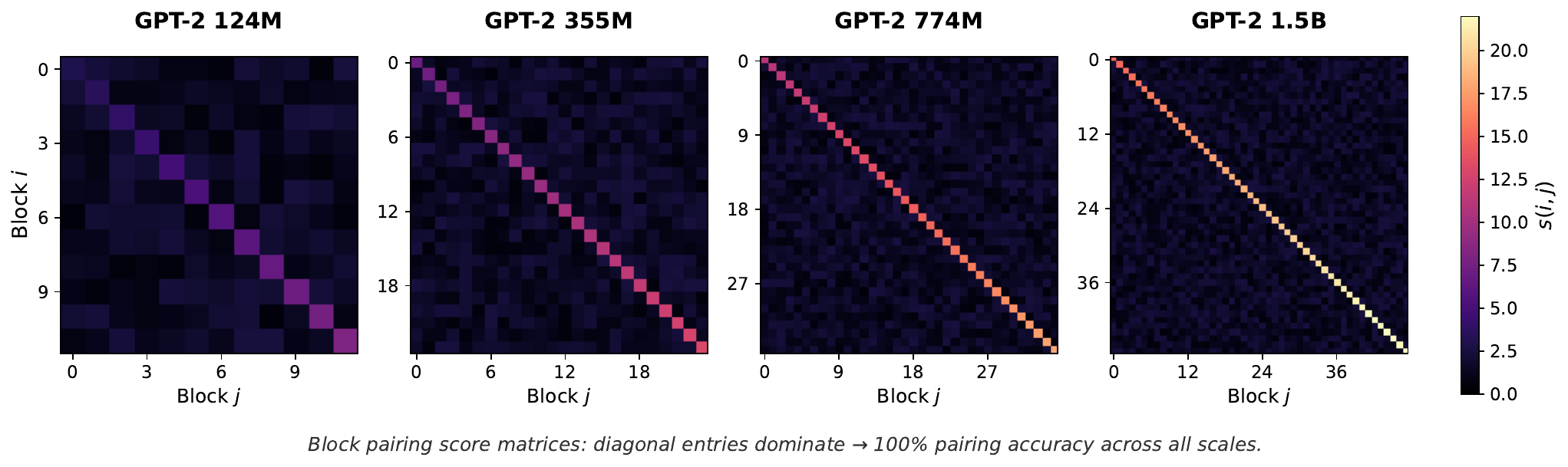}
\caption{Block pairing score matrices $s(i,j)$ for GPT-2 models from 124M to 1.5B parameters. Diagonal entries dominate, yielding 100\% block pairing accuracy across all scales.}
\label{fig:gpt2-pairing}
\end{figure*}

Figure~\ref{fig:gpt2-pairing} shows block pairing score matrices $s(i,j)$ for GPT-2 models from 124M to 1.5B parameters.

\begin{figure*}[t]
\centering
\includegraphics[width=0.95\textwidth]{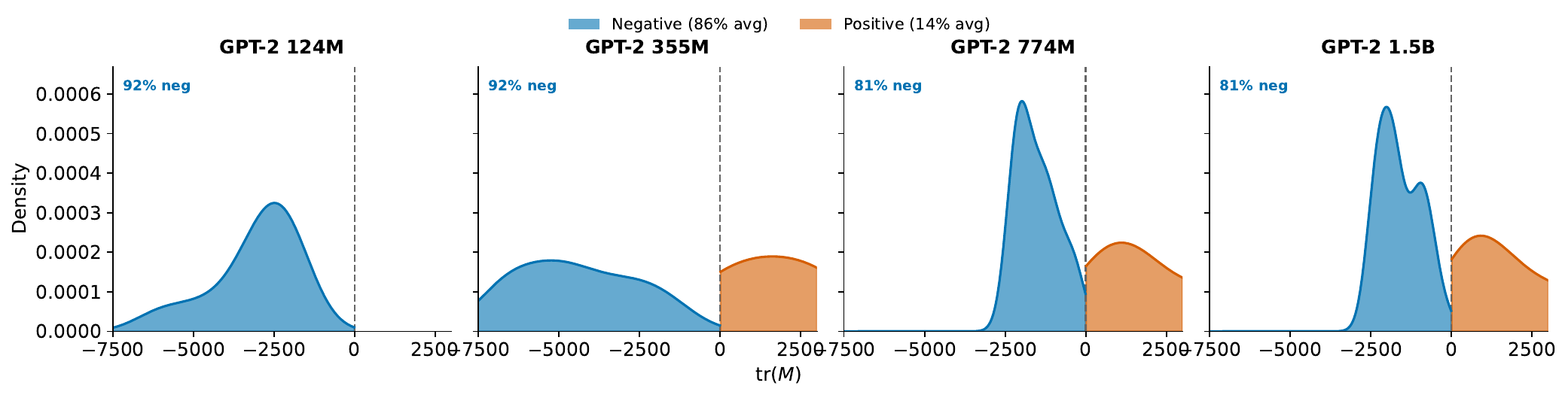}
\caption{Distribution of trace values $\mathrm{tr}(W_{\mathrm{out}} W_{\mathrm{in}})$ across GPT-2 scales. Blue: negative trace (86\% average); orange: positive trace (14\% average). The strong skew toward negative values is consistent with the identity-alignment mechanistic account.}
\label{fig:gpt2-trace}
\end{figure*}

Figure~\ref{fig:gpt2-trace} shows the trace values $\mathrm{tr}(W_{\mathrm{out}} W_{\mathrm{in}})$ per block across all four GPT-2 variants.

\begin{figure}[t]
\centering
\includegraphics[width=0.75\columnwidth]{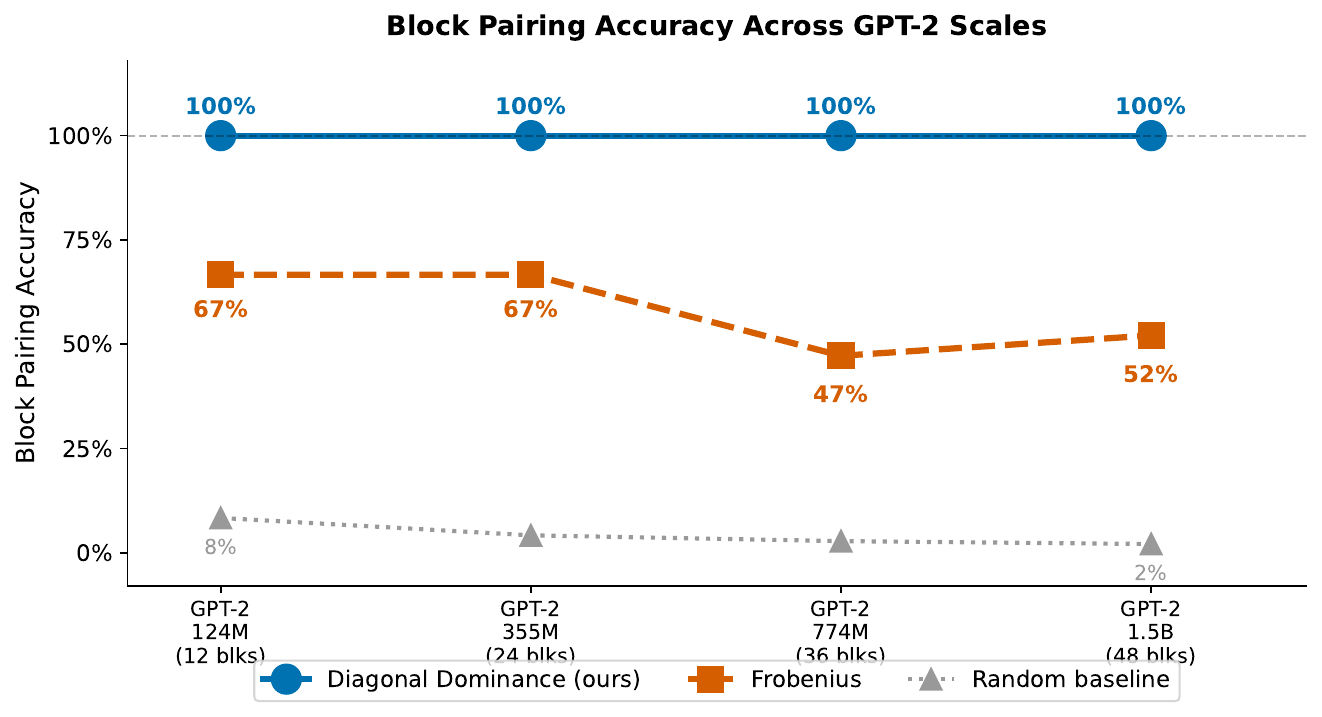}
\caption{Block pairing accuracy as model size increases. Normalized trace pairing (blue) maintains 100\% projection-pair accuracy; Frobenius matching (orange) degrades from 67\% to 52\%; random baseline (gray) drops from 8\% to 2\% as the number of blocks increases.}
\label{fig:gpt2-methods}
\end{figure}

Figure~\ref{fig:gpt2-methods} compares block pairing accuracy across GPT-2 scales.

\subsection{Centering Ablation}\label{app:centering-ablation}

The centering step removes the identity-aligned component shared by all trained models. We verify this materially improves discrimination by directly comparing cosine similarity on uncentered branch products $\cos(\mathrm{vec}(M_\ell^A), \mathrm{vec}(M_\ell^B))$ versus centered residuals $\cos(\mathrm{vec}(R_\ell^A), \mathrm{vec}(R_\ell^B))$ where $R_\ell = M_\ell - \tfrac{\mathrm{tr}(M_\ell)}{d} I$.

Table~\ref{tab:centering-ablation} shows results on the GPT-2 benchmark (27 pairs from 3 test roots). Both methods achieve AUROC$=$1.0, but centering reduces unrelated-pair similarity by ${\sim}50\times$: from 0.019 (uncentered) to 0.0004 (centered). For independent roots specifically, uncentered similarity averages 0.027 while centered similarity averages 0.0015, an $18\times$ reduction.

\begin{table}[h]
\centering
\small
\setlength{\tabcolsep}{3pt}
\begin{tabular}{@{}lcccc@{}}
\toprule
Method & AUC & Gap-$Z$ & Unrel.\ $\mu$ & Margin \\
\midrule
Uncentered $\mathrm{vec}(M_\ell)$ & 1.0 & $+28.7$ & 0.019 & 0.84 \\
Centered $\mathrm{vec}(R_\ell)$ & 1.0 & $+29.2$ & \textbf{0.0004} & \textbf{0.86} \\
\bottomrule
\end{tabular}
\caption{Centering ablation on the GPT-2 benchmark. Both methods perfectly separate related from unrelated pairs, but centering reduces spurious similarity by ${\sim}50\times$.}
\label{tab:centering-ablation}
\end{table}

The identity energy fraction $|\mathrm{tr}(M)|^2 / \|M\|_F^2$ averages 9.4 across all layers, confirming that the identity component carries substantial matrix energy. Independent models share this generic identity-aligned structure from training dynamics; centering removes it, revealing the checkpoint-specific remainder that differs between unrelated models.

\section{Extended Benchmarks}\label{app:benchmarks}

\subsection{Laundering Experiment Gap-Z Scores}\label{app:laundering-gapz}

Table~\ref{tab:laundering-gapz} reports the same laundering experiment as Table~\ref{tab:laundering} but with Gap-$Z$ instead of AUROC. Gap-$Z$ measures the standardized separation between descendant and non-descendant score distributions; larger values indicate more robust margins.

\begin{table}[h]
\centering
\small
\setlength{\tabcolsep}{3pt}
\begin{tabular}{@{}llccccc@{}}
\toprule
& & \multicolumn{5}{c}{Gap-$Z$} \\
\cmidrule(lr){3-7}
Cond. & Bench & \textbf{Ours} & Re-Basin & Al.~Frob & SVD & W.~Cos \\
\midrule
none & MLP & $+53.0$ & $+48.2$ & $+72.0$ & $+7.9$ & $+76.3$ \\
P & MLP & $+53.0$ & $+48.2$ & $-0.1$ & $+7.9$ & $+2.1$ \\
P & GPT-2 & $+31.0$ & $+29.5$ & $+3.9$ & $+0.4$ & $+0.8$ \\
\midrule
D$_\mathrm{m}$ & MLP & $+53.0$ & $+48.2$ & $+72.0$ & $-5.2$ & $+76.3$ \\
D$_\mathrm{s}$ & MLP & $+53.0$ & $+48.2$ & $-4.8$ & $-5.2$ & $+76.3$ \\
PD & MLP & $+53.0$ & $+48.2$ & $-4.8$ & $-5.2$ & $+1.8$ \\
PDFT & MLP & $+49.1$ & $+45.0$ & $-4.8$ & $-5.2$ & $+1.8$ \\
\bottomrule
\end{tabular}
\caption{Gap-$Z$ under function-preserving laundering (same experiment as Table~\ref{tab:laundering}). Weight cosine collapses from $+76.3$ to $+2.1$ under permutation and $+1.8$ under PD; the margin is razor-thin despite AUROC$=$0.80--0.86. Our signature maintains Gap-$Z \approx +53$ across all conditions.}
\label{tab:laundering-gapz}
\end{table}

Weight cosine's Gap-$Z$ collapses from $+76.3$ (none) to $+2.1$ (P) and $+1.8$ (PD). Despite preserving AUROC$=$0.80--0.86, the descendant and non-descendant distributions nearly overlap. One additional perturbation would likely break the ranking entirely. Our signature maintains Gap-$Z \approx +53$ across all conditions, indicating stable separation regardless of laundering.

\subsection{Harder Regime: Layer Grafts and Linear Merges}\label{app:harder-bench}

We construct 40 harder pairs from depth-16 MLPs (same setup as Table~\ref{tab:baselines}): (i) \textbf{layer grafts} copy $K \in \{0,4,8,12,16\}$ of 16 blocks from reference into an independent donor (related if $K \ge 8$); (ii) \textbf{linear merges} compute $w \cdot \mathrm{ref} + (1-w) \cdot \mathrm{donor}$ with $w \in \{0, 0.25, 0.5, 0.75, 1.0\}$ (related if $w \ge 0.5$). We use 2 references and 2 donors per reference. AUROC measures binary related-vs-unrelated detection; Spearman correlation measures whether a method tracks partial overlap monotonically ($K/L$ or $w$) rather than merely clearing a threshold. Table~\ref{tab:harder-bench} summarizes the 40-pair benchmark.

\begin{table}[h]
\centering\small
\setlength{\tabcolsep}{2pt}
\begin{tabular}{@{}lcccc@{}}
\toprule
& \multicolumn{2}{c}{AUROC} & \multicolumn{2}{c}{Spearman $\rho$} \\
\cmidrule(lr){2-3}\cmidrule(lr){4-5}
Method & Graft & Merge & Graft & Merge \\
\midrule
Ours  & 0.98 & 1.00 & +.96 & +.96 \\
Aligned Frob.\ & 1.00 & 1.00 & +.99 & +.99 \\
Weight cos.\ & 1.00 & 1.00 & +.98 & +.96 \\
SVD dist.\ & 1.00 & 0.45 & +.99 & +.23 \\
SVCCA & 1.00 & 1.00 & +.97 & +.99 \\
CKA & 0.81 & 0.78 & +.73 & +.75 \\
IPGuard & 0.05 & 0.45 & $-$.83 & +.22 \\
\bottomrule
\end{tabular}
\caption{Harder-regime benchmark on layer grafts and linear merges (40 pairs). Our method tracks partial overlap monotonically ($\rho \ge 0.96$). CKA degrades; IPGuard collapses on grafts; SVD fails merges.}
\label{tab:harder-bench}
\end{table}

\subsection{CIFAR-10 ResNet-18 Benchmark}\label{app:cifar-resnet}

We train two CIFAR-10 ResNet-18 references and three independents. Each reference yields 11 related checkpoints (noise $1{-}10\%$, pruning $20{-}80\%$, quantization $16{-}256$ levels). Results: AUROC$=$1.000; all related checkpoints exceed the maximum null score. Related checkpoint scores: $\mathcal{L} \in [0.695, 1.000]$; unrelated baseline: $\mathcal{L}_{\max} = 0.004$.

\subsection{GPT-2-Small-Lite Benchmark}\label{app:gpt2-benchmark}

\begin{figure}[h]
\centering
\includegraphics[width=\columnwidth]{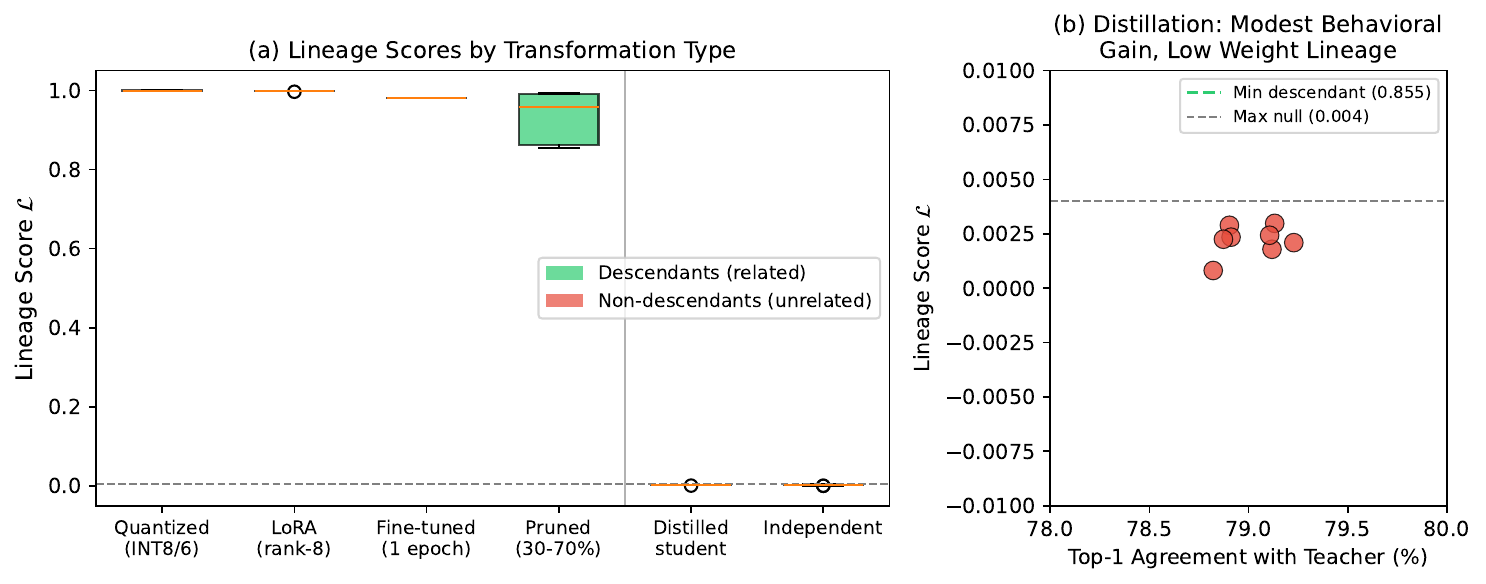}
\caption{GPT-2-Small-Lite lineage benchmark (8 roots, 120 pairs across all splits; Table~\ref{tab:gpt2-survival} reports results on the 3 held-out test roots, 45 pairs). (a) Lineage scores by transformation type. Descendants (green) score high; non-descendants (red) score near zero. (b) Distillation modestly improves teacher--student agreement (79\% top-1, +1.4 points over independent training) while leaving the weight-lineage score near the independent-root range, confirming that behavioral similarity $\neq$ weight inheritance.}
\label{fig:gpt2-benchmark}
\end{figure}

Figure~\ref{fig:gpt2-benchmark} visualizes the GPT-2-Small-Lite benchmark results from Section~\ref{sec:exp-posttraining}. Panel (a) shows lineage score distributions by transformation type: descendants (quantized, LoRA, fine-tuned, pruned) score high while non-descendants (distilled students, independent roots) score near zero. The minimum descendant score (0.855) exceeds the maximum non-descendant score (0.004). Panel (b) shows that distillation modestly improves teacher--student agreement (79\% top-1, +1.4 points over independent training) yet leaves the weight-lineage score near the independent-root range ($\mathcal{L} \approx 0.002$). This illustrates the conceptual distinction: behavioral similarity does not imply weight ancestry.

\subsection{ROC for Lineage Verification}

\begin{figure}[h]
\centering
\includegraphics[width=0.85\columnwidth]{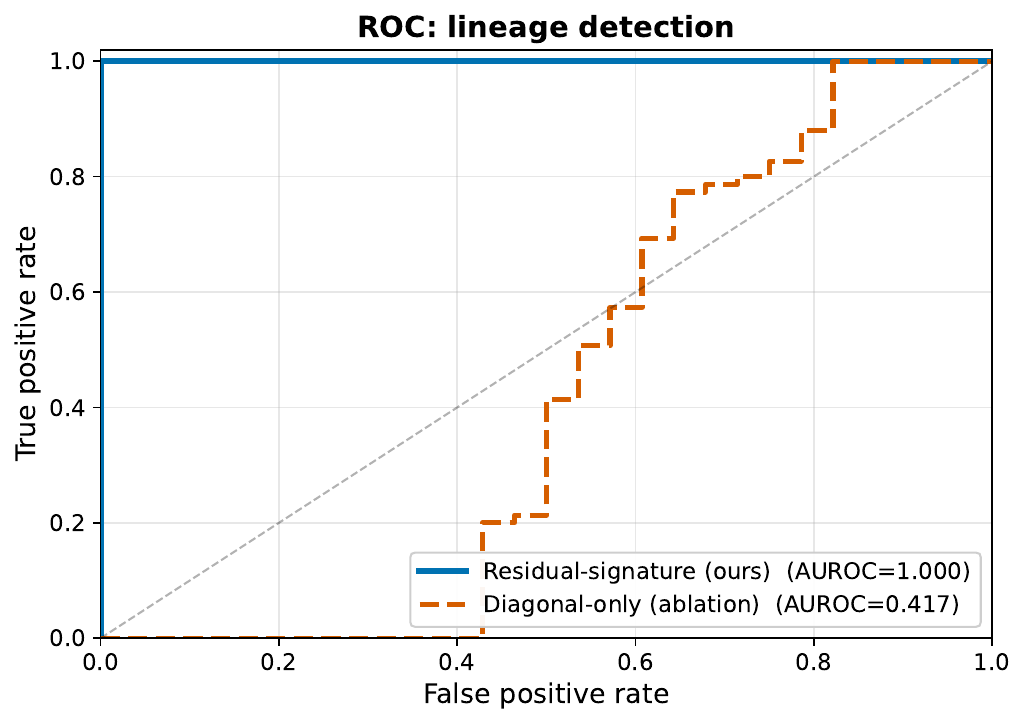}
\caption{ROC for lineage verification on depth-24 residual MLPs. The residual-signature score achieves AUROC$=$1.000; trace concentration alone fails (AUROC$=$0.417) because it cannot distinguish two trained residual models.}
\label{fig:lineage-roc}
\end{figure}

Figure~\ref{fig:lineage-roc} shows that trace concentration alone cannot distinguish related from unrelated trained models.

\subsection{Gradient-Based Suppression Attack}\label{app:attacks}

An adversary optimizes perturbation $\Delta$ to minimize lineage score subject to utility:
\begin{equation}
\min_{\Delta} \;\mathcal{L}_{\mathrm{cos}}(A, A + \Delta) \;+\; \lambda \cdot \|f_{A+\Delta}(X) - f_A(X)\|_2^2
\end{equation}

We run this attack on depth-24 MLPs ($d{=}64$). The null baseline for this configuration (score between the reference and an independent same-task model) is $\mathcal{L}_{\text{null}} = 0.084$.

\begin{table}[h]
\centering
\small
\begin{tabular}{@{}rccl@{}}
\toprule
$\lambda$ & Final $\mathcal{L}$ & Eval loss & Verdict \\
\midrule
$0$         & 0.053 & 39.5 & utility destroyed \\
$10^{-2}$   & 0.054 & 0.77 & +12\% loss, below null \\
$10^{-1}$   & 0.083 & 0.70 & +1.5\% loss, $\approx$ null \\
$\ge 1$     & 0.91+ & 0.69 & utility preserved \\
\bottomrule
\end{tabular}
\caption{Pareto frontier for suppression attack. The null baseline for this setup is $\mathcal{L}_{\text{null}} = 0.084$. Reaching the null threshold costs +1.5\% utility loss ($\lambda{=}10^{-1}$); driving the score reliably below null ($\lambda{=}10^{-2}$) costs +12\%.}
\label{tab:attack-pareto}
\end{table}

\section{Transfer Beyond Language Models}\label{app:beyond-lm}

The trace concentration phenomenon underlying our method is not specific to language models. Table~\ref{tab:vision-speech} shows that architecture-aware branch products recover projection pairings in vision and speech architectures as well.

\begin{table}[h]
\centering
\footnotesize
\setlength{\tabcolsep}{4pt}
\begin{tabular}{@{}llcc@{}}
\toprule
Model & $L$ & Pair Acc & AUC \\
\midrule
ViT-B/16 & 12 & \textbf{100\%} & 1.00 \\
Whisper (tiny/base/sm) & 4--12 & \textbf{100\%} & 0.85 \\
ResNet-50/101/152 & 5--35 & 91--100\% & 0.96 \\
\bottomrule
\end{tabular}
\caption{Trace concentration in vision and speech architectures. Pair accuracy measures correct within-block projection recovery via Hungarian matching on $s(i,j)$. Random-init baselines: $\le 9\%$.}
\label{tab:vision-speech}
\end{table}

For ViT~\citep{dosovitskiy2020vit}, we use the same transformer MLP factorization as for language models ($W_2 W_1$). For Whisper~\citep{radford2023whisper}, we extract branch products from both encoder and decoder MLP blocks. For ResNet~\citep{he2016resnet}, bottleneck blocks require a three-layer product ($W_3 W_2 W_1$); naive two-layer factorization ($W_3 W_1$, skipping the middle conv) yields chance-level accuracy. On ImageNet-pretrained ResNet-50/101/152 (torchvision), the correct triple product recovers 91--100\% of block pairings in layer3.

\end{document}